\documentclass[sigconf]{acmart}
\usepackage[ruled,vlined]{algorithm2e}
\usepackage[shortlabels]{enumitem}
\AtBeginDocument{
  \providecommand\BibTeX{{
    \normalfont B\kern-0.5em{\scshape i\kern-0.25em b}\kern-0.8em\TeX}}}

\acmConference[AdvML '21]{AdvML '21: Workshop on Adversarial Learning Methods for Machine Learning and Data Mining}{August 15, 2021}{Singapore (virtual)}
\acmBooktitle{AdvML '21: Workshop on Adversarial Learning Methods for Machine Learning and Data Mining, August 15, 2021, Singapore (virtual)}

\begin{document}

\title{Detection and Recovery Against Deep Neural Network Fault Injection Attacks Based on Contrastive Learning}

\author{Chenan Wang, Pu Zhao, Siyue Wang, Xue Lin}
\affiliation{
    \institution{
    Northeastern University, Boston, MA, USA
    }
    \country{}
}
\email{{wang.chena, zhao.pu, wang.siy, xue.lin}@northeastern.edu}

\renewcommand{\shortauthors}{Wang, Zhao, et al.}

\begin{abstract}
Deep Neural Network (DNN) models when implemented on executing devices as the inference engines are susceptible to Fault Injection Attacks (FIAs) that manipulate model parameters to disrupt inference execution with disastrous performance. This work introduces Contrastive Learning (CL) of visual representations i.e., a self-supervised learning approach into the deep learning training and inference pipeline to implement DNN inference engines with self-resilience under FIAs. Our proposed CL based FIA Detection and Recovery (CFDR) framework features (i) real-time detection with only a single batch of testing data and (ii) fast recovery effective even with only a small amount of unlabeled testing data. Evaluated with the CIFAR-10 dataset on multiple types of FIAs, our CFDR shows promising detection and recovery effectiveness.

\end{abstract}



\maketitle
\section{Introduction}
Deep learning (DL) has succeeded in many application domains such as computer vision~\cite{he2016deep,szegedy2017inception} and   natural language processing~\cite{hinton2012deep,devlin2018bert}.
Along with the prosperity of DL, its vulnerability under adversarial attacks has drawn significant attentions.
For example, sophisticatedly crafted perturbations can be added onto clean images to produce adversarial examples \cite{szegedy2014intriguing,xu2019structured}, the prediction of which by Deep Neural Networks (DNNs) will be erroneous, although the added perturbations are mostly imperceptible to humans. 

Besides adversarial  examples, 
Fault Injection Attacks (FIAs) present another category of adversarial attacks, which aim at DNN inference models when implemented on executing devices.
In general, FIAs modify DNN model parameters, leading to malfunction of the inference models e.g., severely degradation in prediction accuracy or targeted misclassification of specified objects. Therefore, a DNN inference engine is subject to integrity violation caused by FIAs.

There are several state-of-the-art DNN FIAs proposed with diverse  algorithms.
Liu et al. proposed the first FIA with a heuristic algorithm i.e., Gradient Descent Attack (GDA) \cite{fia_dnn} that modifies DNN model parameters to classify specified inputs into wrong labels.
Furthermore, Fault Sneaking Attack (FSA) \cite{fsa} improved upon GDA with ADMM (Alternating Direction Method of Multipliers) based algorithms that set two constraints i.e., $\ell_0$ or $\ell_2$ norm of the parameter modifications, besides the specified misclassifications.
On the other hand, He, Rakin, et al. proposed a FIA for quantized DNN models i.e., Bit Flip Attack (BFA) \cite{bfa}, which randomly picks model parameters and selects the most sensitive bit to flip.
Then Progressive Bit Search (PBS) \cite{pbs} extended BFA with cross-layer and intra-layer searches.
Please note that both BFA and PBS target for tampering with DNN models through minimal modification efforts.

This paper aims to design DNN models with self-resilience under FIAs.
We are the first to use Contrastive Learning (CL) towards this objective.
Specifically, we use CL to obtain the DNN inference models. Comparing with conventional DNN training, CL enables our detection as well as recovery mechanisms against FIAs.
With DNN inference models obtained by CL, we propose to observe the change of contrastive loss over single batches of testing data to detect potential FIAs.
Since the detection mechanism does not require labeled data, it can be executed periodically without disruption on the normal DNN inference process.
Whenever a FIA is detected, the recovery algorithm will be triggered to boost the accuracy performance close to that before the FIA.
We consider two scenarios for the recovery process i.e., the labeled training data is available or only unlabeled testing data is available for recovery.
In both scenarios, our proposed recovery algorithm can boost the accuracy significantly with only a small amount of data in a few  number of epochs.
We summarize our contributions as follows:
\begin{itemize}
    \item The first CL based approach for FIA detection and recovery.
    \item A highly sensitive detection   requiring a single batch of unlabeled   data without disruption on   normal inference process.
    \item A fast recovery algorithm that significantly boosts accuracy, even with only a small amount of unlabeled data.
\end{itemize}

\section{Proposed CL based FIA Detection and Recovery (CFDR)}

This section introduces our main approaches by first fitting CL for our purpose, and then presenting our proposed detection and recovery mechanisms.

\subsection{Contrastive Learning and Preliminaries}
Contrastive learning has been proposed as a self-supervised learning to reduce the requirement on the amount of labeled training data.
For contrastive learning, we adopt the SimCLR  method \cite{simclr} that learns representations by maximizing agreement between differently augmented views of the same data example via a contrastive loss in the latent space. 
We will find in the following that (i) the contrastive loss is a key criterion in our FIA detection, and (ii) the relaxed requirement on the amount of labeled data by CL enables our effective recovery from FIA.

\begin{figure}[htb!]
  \centering
  \centerline{\includegraphics[width=9cm]{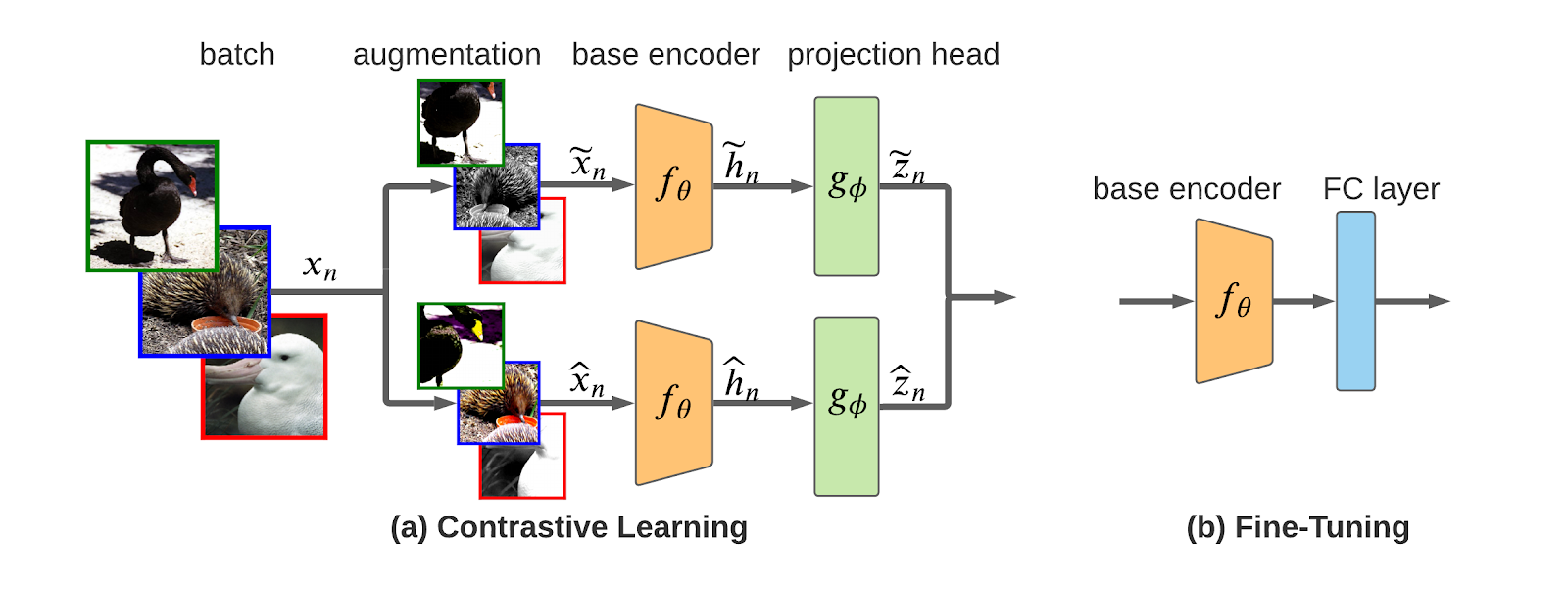}}
\caption{The SimCLR framework.}
\label{fig: detection_fig}
\end{figure}

Figure \ref{fig: detection_fig} (a) and (b) present the two phases in the whole contrastive learning pipeline. We will first introduce the four components for Phase (a) Contrastive Learning:
\begin{itemize}
    \item \textbf{A stochastic data augmentation module.} The module performs  random combinations of data augmentation methods for each original data example $x_n$ to generate a pair of  correlated views of the same example, denoted by $\widehat{x}_n $ and $\widetilde{x}_n$,  which are considered as a positive pair. 
    \item \textbf{A base encoder $f_\theta$.} It extracts image representation/embedding from augmented data examples. 
    Following the SimCLR method, we also use ResNet (without the last fully-connected (FC) layer) as the base encoder to have
\begin{equation}
    h_n=f_\theta (x_n).
\end{equation}
    \item \textbf{A projection head $g_\phi$.} It projects/maps high dimensional image representations to latent space  with lower dimensionality, where the contrastive loss can be applied. It is a shallow multi-layer perceptron (MLP) with one hidden layer i.e., 
    \begin{equation}
    z_n=g_\phi(h_n).
\end{equation}
    \item \textbf{A contrastive loss function.} It is designed for maximizing agreement between both image embeddings of the same data examples i.e., the positive pairs as follows: 
    \begin{equation}
   {Loss}=-\sum_{n=1}^{N} \log \frac{\exp(\text{sim}(\widetilde{\mathbf{z}}_{n},\widehat{\mathbf{z}}_{n}) / \tau)}{\sum_{k=1, k\neq n}^{N}\exp(\text{sim}(\widetilde{\mathbf{z}}_{n},\widehat{\mathbf{z}}_{k})/\tau)+\exp(\text{sim}(\widehat{\mathbf{z}}_{n},\widetilde{\mathbf{z}}_{k})/\tau)}  
\end{equation}
where $(\widetilde{\mathbf{z}}_{n},\widehat{\mathbf{z}}_{n})$ represents a positive pair, and $(\widetilde{\mathbf{z}}_{n},\widehat{\mathbf{z}}_{k})$ or $(\widehat{\mathbf{z}}_{n},\widetilde{\mathbf{z}}_{k})$ represents a negative pair. The $\text{sim}(\cdot)$ function is the  dot product between  two $\ell_2$ normalized  vectors (i.e., cosine similarity).
\end{itemize}

With those above described components, the two phases in the whole contrastive learning pipeline are summarized as below:
\begin{enumerate}[label=(\alph*)]
    \item \textbf{Contrastive Learning Phase} as in Figure \ref{fig: detection_fig} (a). It trains the base encoder and the projection head with the contrastive loss. Only unlabeled data is needed in this phase. 
    \item \textbf{Fine-Tuning Phase} as in Figure \ref{fig: detection_fig} (b). It fine-tunes on the FC layer using the regular cross entropy loss with a small amount of labeled training data, while the base encoder  that is already trained in Phase (a) is not updated. Note that in our paper the base encoder plus the FC layer make a whole ResNet model architecture. 
    The architecture in Figure \ref{fig: detection_fig} (b) is the finally obtained DNN inference model i.e., a SimCLR trained ResNet model.
\end{enumerate}

\subsection{Detection and Recovery Overview}

With a DNN model obtained through the SimCLR method introduced in the previous section, now we introduce the overview of our proposed CL based FIA Detection and Recovery (CFDR) framework.

For the detection of FIAs, we propose to use the contrastive learning loss in Phase (a) as a key criterion to determine whether the DNN model is attacked by a FIA.
We use the contrastive learning loss over a batch of data calculated with the clean (unattacked) model as the reference value.
Note that the contrastive loss does not rely on labeled data, and therefore, the detection process can be co-executed during the normal DNN inference. 
Once a higher contrastive loss over a batch of testing data is observed, we determine that a FIA was conducted on the model.
In summary, our detection mechanism features the real-time detection without interference with the normal inference execution.

Once a FIA is detected, our recovery algorithm will be triggered to boost the accuracy close to that before the FIA.
We consider two scenarios for the recovery process.
If the labeled training data is available for the recovery, we can recover a fault injection attacked model wherever the FIA is conducted on i.e., any layers of the DNN model.
If only unlabeled testing data is available for the recovery, our recovery process can only address the cases where the FIA is conducted on any layers except the last FC layer.
This is the limitation of our recovery mechanism.

\subsection{Detection Mechanism}

The detection of FIA is co-executed with the normal inference process with unlabeled testing data.
Specifically, the contrastive loss over a batch of testing data is calculated with the same architecture as Figure \ref{fig: detection_fig} (a).
Although this architecture is not directly used during inference, we can add the project head at the output of base encoder as the other path in parallel with the FC layer. This newly added path is for calculating contrastive loss simultaneouly with the inference execution.
Therefore, we can implement real-time detection by comparing the contrastive loss with that of the clean model i.e., the reference value.
If a relative large difference is observed, then the model is attacked.
Besides the reference value, we set the fault tolerance parameter as the threshold for the difference.
This fault tolerance parameter should be properly set to reduce the occurrences of false positives and false negatives.
Furthermore, we can also use the contrastive loss over multiple batches for more accurate detection.

\begin{algorithm}[t]
\SetAlgoLined
\KwData{The model to be detected, unlabeled testing data for detection, and a small amount of unlabeled/labeled data for recovery.}
\KwResult{The detection result and the recovered model if a FIA is detected.}
Train a new DNN with the two-phase SimCLR method\;
Compute the average contrastive loss $l_c$ for the clean model\;
Set $l_c$ of the clean model as the reference value\;
{Set $\delta$ as the fault tolerance parameter\;}
Compute the contrastive loss $l_d$ for the  model to be detected over a batch of unlabeled testing data\;
\eIf{$|l_d - l_c| > \delta$}{
   the model is attacked\;
   perform model recovery\\
   \eIf{labeled data is available}{
   perform contrastive learning phase (a) for the model\;
   perform fine-tuning phase (b)\;
   }{ 
   perform contrastive learning phase (a)\;
   }
   }{
   the model is not attacked\;}
 \caption{CL based FIA Detection and Recovery.}
 \label{alg:detect}
\end{algorithm}

\subsection{Recovery Mechanism}

Once a FIA is detected, the recovery process is needed to boost the accuracy performance by retraining the model with a small amount labeled or unlabeled data.
For the case that labeled training data is available, we perform both Phase (a) and Phase (b) for multiple epochs over the available training data.
For the case that only unlabeled testing data is available for recovery, we perform only Phase (a).

Due to the small amount of labeled/unlabeled data for recovery, cautions must be used to avoid overfitting. 
Therefore, we use the following stopping criteria:
\begin{enumerate}
    \item The training loss is less than or equal to a reference value. 
    For the contrastive loss in Phase (a) or the cross entropy loss in Phase (b), we use their counterpart on the clean model as the reference value. \textbf{OR}
    \item The training loss stops decreasing. \textbf{OR}
    \item The total epoch number reaches a certain value e.g., 30. 
\end{enumerate}

\section{Experiments}
\subsection{Experimental Setup}

We evaluate our CFDR framework using a ResNet-18 model trained with the SimCLR framework on CIFAR-10 dataset.
We adopt a batch size of 64 throughout the whole process i.e., SimCLR training, detection and recovery.
For the original SimCLR training, Phase (a) uses 1,000 epochs and Phase (b) uses 100 epochs, i.e., the same setting as the SimCLR codes.

For detection, the contrasive loss over a single batch of testing data is sampled 1,000 times to obtain the detection results in Section \ref{sec:detectresult}.
For recovery, we assume only 512 images are available for both the unlabeled and labeled data cases.

For evaluation, we adopt four types of FIAs i.e., PBS \cite{pbs}, FSA $\ell_0$ \cite{fsa}, FSA $\ell_2$ \cite{fsa}, and GDA \cite{fia_dnn}.
We use  attacked models by the above-mentioned four attacks in various settings to evaluate our CFDR framework.

\subsubsection{PBS}
Progressive bit search (PBS) \cite{pbs} performs in-layer search and cross-layer search.
In-layer search finds the most vulnerable bits from a layer; cross-layer search finds the most vulnerable layer with in-layer search. The goal of PBS is to tamper with the network, more precisely, degrading the top-1 accuracy of attacked network below 11\%.
For PBS, we adopt the default setting  of \cite{pbs}, i.e. the hacker can change all parameters in the attacked layer. Normally, single PBS run tampers with the network below 11\%.

\subsubsection{FSA}
Fault sneaking attack (FSA) \cite{fsa} uses efficient ADMM (alternating direction method of multipliers) algorithms to  modify model parameters, so that the model would make wrong predictions.
In the experiment, we found out that FSA usually changes almost all parameters in a layer. 
For FSA, we set S=5 and R=20, i.e. we modify 5 images out of 20 images. 5 images are misclassified to wrong labels while 15 images keeps the original correct labels.

\subsubsection{GDA}
Similar to FSA, Gradient descent attack (GDA) \cite{fia_dnn} is more straightforward.
It  gradually modifies parameters with gradient information to enlarge the predictions of a specific class, leading to incorrect predictions. 
At the same time, they use an $\ell_1$-norm regulator to limit the parameter modifications. 
Different from the original setting, we do not perform modification compression, since its iteration is inefficient. Besides, we use an $\ell_2$ regulator to restrict model modifications.

\subsection{Results on Detection}\label{sec:detectresult}
\begin{figure}[h]
  \centering
  \includegraphics[width=0.81\linewidth]{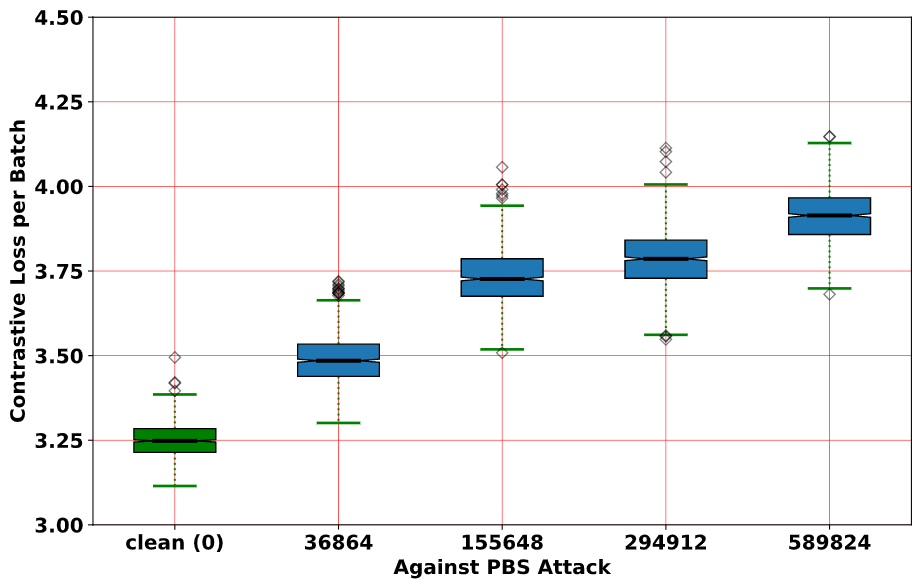}
  \caption{Detection effectiveness by box plot when different number of parameters are modified by the PBS}
  \label{fig:detect_PBS}
\end{figure}

\begin{figure}[h]
  \centering
  \includegraphics[width=0.81\linewidth]{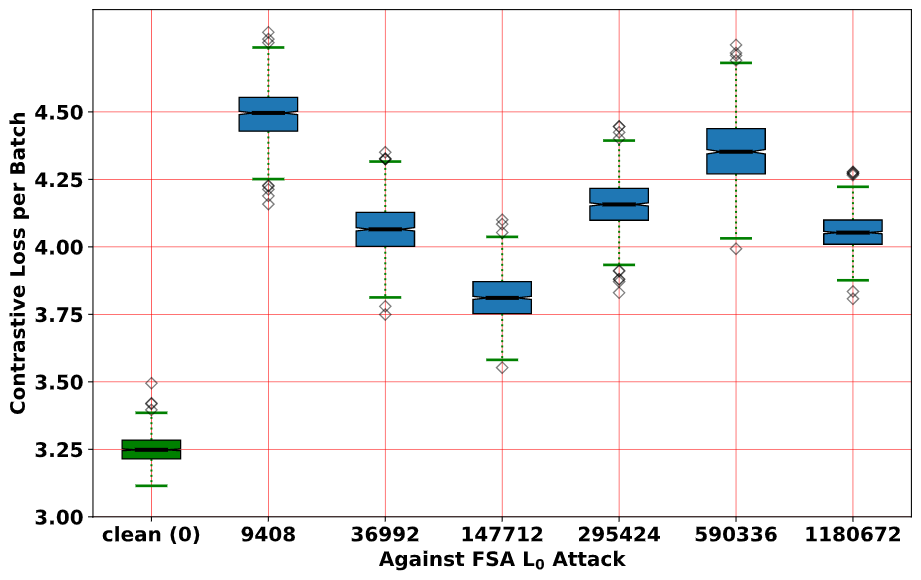}
  \caption{Detection effectiveness by box plot when different number of parameters are modified by the FSA $\ell_0$}
  \label{fig:detect_FSAl0}
\end{figure}

\begin{figure}[h]
  \centering
  \includegraphics[width=0.81\linewidth]{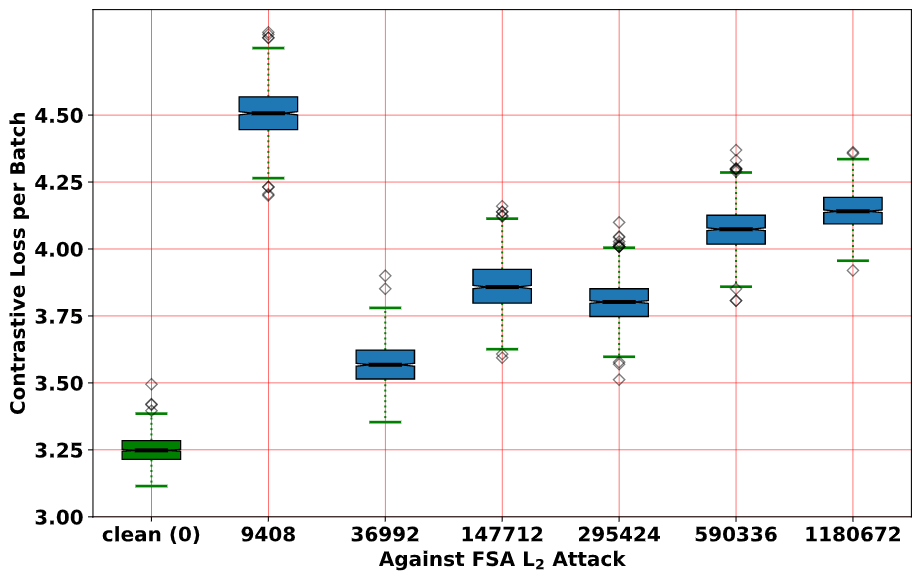}
  \caption{Detection effectiveness by box plot when different number of parameters are modified by the FSA $\ell_2$}
  \label{fig:detect_FSAl2}
\end{figure}

\begin{figure}[h]
  \centering
  \includegraphics[width=0.81\linewidth]{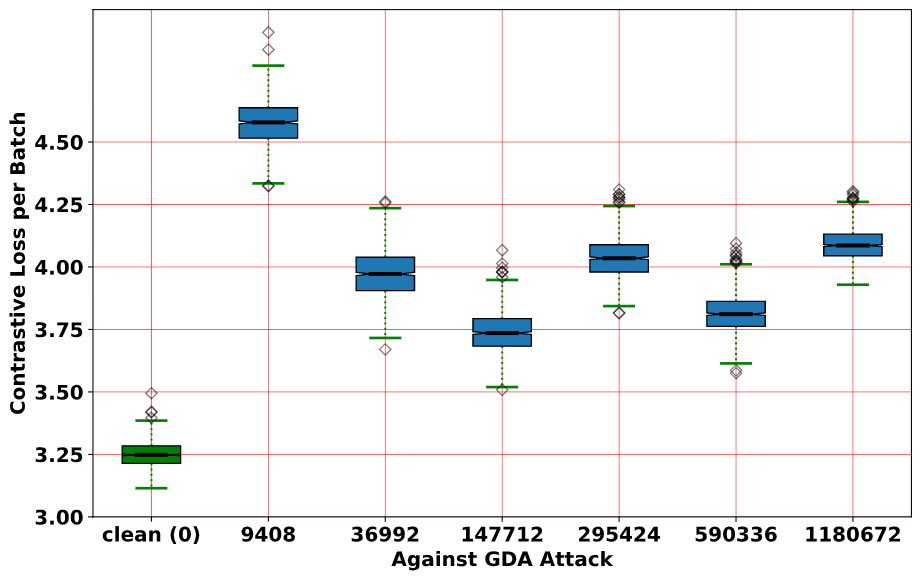}
  \caption{Detection effectiveness by box plot when different number of parameters are modified by the GDA}
  \label{fig:detect_GDA}
\end{figure}

Figures \ref{fig:detect_PBS}, \ref{fig:detect_FSAl0},  \ref{fig:detect_FSAl2}, and \ref{fig:detect_GDA} represent the detection effectiveness of our CFDR framework against PBS, FSA $\ell_0$, FSA $\ell_2$, and GDA attacks.
For each attack type, we conducted FIA multiple times, acting on different layers of the DNN model.
The number of parameters in the modified layers of each attack instance is the x-axis.
The contrastive loss is sampled 1,000 times over single batches of testing data.
As can be observed from the figures, the contrastive loss is well seperated from that of the clean model, demonstrating the effectiveness of our detection mechanism. 

\setlength{\tabcolsep}{3pt}

\subsection{Results on Recovery}

Table \ref{table_mnist_lottery} presents the recovery effectiveness of our CFDR framework

For each FIA type, we generate various attacked models by perturbing different layers with different number of parameters. As shown in Table \ref{table_mnist_lottery}, the accuracy after attack suffers from obvious degradation. 
But after we successfully detected the attacks, we can perform model recovery with labeled or unlabeled data. After the recovery, we are able to improve the accuracy on the test set. For models with less perturbed/attacked parameters, the accuracy   can be restored to a typical accuracy (around 87\%) for ResNet-18 on CIFAR-10 with only a few data. For models with more perturbed/attacked parameters (e.g., 1180672 parameters), the accuracy can  be improved to 45\% for GDA with labeled data, which is still smaller than the normal accuracy with a large gap. This demonstrates the limitations/difficulties for recovery  to restore if too many parameters (e.g. more than 1 billion) are perturbed. We also notice that the recovery with labeled data can usually achieve higher accuracy than the recovery with unlabeled data, demonstrating that more information with labels can help with the training.

\begin{table}[htb!]
\small
 \caption{Recovery effectiveness} 
  \label{table_mnist_lottery}
 \centering
\begin{tabular}{c|c c| c c| c c}
\toprule[1pt]
Attack &\shortstack{total \# of \\param. in \\the attack\\-ed layer(s)} & \shortstack{after\\attack \\ acc.} & \shortstack{unlabeled \\recovery\\acc.}& epochs& \shortstack{labeled \\recovery\\acc.}  & epochs \\
\midrule[1pt]

PBS          &36864 &10.00&88.71&6&88.89&7  \\
PBS          &155648&10.16&87.72&15&87.64&22  \\ 
PBS          &294612& 13.25&87.36&15&87.34&5  \\
PBS          &589824& 10.00 & 80.34 &  12&81.05&5  \\
\hline
FSA $\ell_0$ &9408&12.57&85.15&30&85.57&9\\
FSA $\ell_0$ &36992&10.07&88.09&20&88.11&15\\
FSA $\ell_0$ &147712&10.47&87.68&24&88.22&7\\
FSA $\ell_0$ &295424&14.43&79.63&18&80.82&5\\
FSA $\ell_0$ &590336&17.03&65.84&15&68.52&10\\
FSA $\ell_0$ &1180672&11.86&33.45&21&40.41&10\\
\hline

FSA $\ell_2$ & 9408&11.17&85.23&30&85.65&9\\
FSA $\ell_2$ &36992&13.6&88.50&10&88.41&3\\
FSA $\ell_2$ &147712&11.04&87.53&22&87.93&5\\
FSA $\ell_2$ &295424&15.22&83.70&11&84.19&5\\
FSA $\ell_2$ &590336&10.95&70.57&20&72.37&5\\
FSA $\ell_2$ &1180672&12.84&28.42&19&35.07&15\\
\hline

GDA&9408&46.79&82.77&30&82.93&9\\
GDA&36992&83.42&88.08&9&88.33&9\\
GDA&147712&84.72&87.62&9&87.87&24\\
GDA&295424&55.89&80.41&17&80.86&9\\
GDA&590336&58.69&79.38&27&79.92&10\\
GDA&1180672&25.41&32.79&30&45.36&30\\
\bottomrule[1pt]
\end{tabular}
\end{table}

\section{Conclusion}
This work introduces Contrastive Learning (CL) of visual representations
into DL training and inference pipeline to implement DNN inference engines with self-resilience under FIAs. Our proposed CL based FIA Detection and Recovery (CFDR) framework features (i) the first CL based approach for FIA detection and recovery; (ii) a highly sensitive detection mechanism requiring a single batch of unlabeled testing data without disruption on the normal inference process; and (iii) a fast recovery algorithm that significantly boosts accuracy performance even only with unlabeled testing data.
Evaluated with the CIFAR-10 dataset on multiple types of FIAs, our CFDR shows promising detection and recovery effectivenesses.

\begin{acks}
This work is supported by National Science Foundation (NSF) under Award \# CNS-1929300.

\end{acks}

\bibliographystyle{ACM-Reference-Format}
\bibliography{reference}

\end{document}